%% file: main.tex
\documentclass{article}
\include{notation}
\usepackage{spconf,amsmath,graphicx}
\usepackage{tikz}
\usepackage{pgfplots}
\pgfplotsset{plots/.style= {grid=major,font=\small, height=2.4in, width=\columnwidth}} 

\title{Hybrid multi-layer Deep CNN/Aggregator feature for image classification}

\name{Praveen Kulkarni$^1$, Joaquin Zepeda$^1$, Frederic Jurie$^2$, Patrick Perez$^1$ and Louis Chevallier$^1$\thanks{This work was partially supported by the FP7 European integrated project AXES.}}
\address{$^1$Technicolor 
975 avenue des Champs Blancs,
CS 17616, 35576 Cesson S{\'e}vign{\'e}, France \\
$^2$University of Caen Basse-Normandie, 
 CNRS UMR 6072, ENSICAEN, France}
\ninept
\begin{document}


%
\maketitle
\begin{abstract}
Deep Convolutional Neural Networks (DCNN) have established a remarkable performance benchmark in the field of image classification, displacing classical approaches based on hand-tailored aggregations of local descriptors. Yet DCNNs impose high computational burdens both at training and at testing time, and training them requires collecting and annotating large amounts of training data. Supervised adaptation methods have been proposed in the literature that partially re-learn a transferred DCNN structure from a new target dataset. Yet these require expensive bounding-box annotations and are still computationally expensive to learn.  In this paper, we address these shortcomings of DCNN adaptation schemes by proposing a hybrid approach that combines conventional, unsupervised aggregators such as Bag-of-Words (BoW), with the DCNN pipeline by treating the output of intermediate layers as densely extracted local descriptors.

We test a variant of our approach that uses only intermediate DCNN layers on the standard PASCAL VOC 2007 dataset and show performance significantly higher than the standard BoW model and comparable to Fisher vector aggregation but with a feature that is $150$ times smaller. A second variant of our approach that includes the fully connected DCNN layers significantly outperforms Fisher vector schemes and performs comparably to DCNN approaches adapted to Pascal VOC 2007, yet at only a small fraction of the training and testing cost.
\end{abstract}

%

\section{Introduction}
In this paper we propose a new hybrid image feature for image classification obtained from a mix of the classical image feature extraction pipeline and the more recent and very successful Deep Convolutional Neural Network (DCNN) pipeline. 

The classical image feature extraction pipeline consist of three major steps: 1) Extracting local descriptors such as SIFT \cite{siftdescriptor} from the image; 2) mapping these descriptors to a higher dimensional space; 3) and sum or max-pooling the resulting vectors to form a fixed-dimensional image feature representation. Examples of methods corresponding to this classical approach include Bag-of-Words (BoW) \cite{csurka2004visual}, Fisher Vector (FV) \cite{perron}, Locality-constrained Linear Encoding \cite{Wang2010}, Kernel codebooks \cite{van2008kernel}, super-vector encoding \cite{zhou2010image} and VLAD \cite{Delhumeau2013}. We refer to these type of image feature extraction schemes as \emph{aggregators} given that they aggregate local descriptors into a fixed dimensional representation. Generally these approaches require computationally inexpensive unsupervised models of the local descriptor distribution, and the resulting image features can be used to learn likewise inexpensive linear classifiers using SVMs. 

The novel DCNN  pipeline of \cite{Krizhevsky2012} has drastically pushed the performance limits of image classification. DCNNs consist of multiple interconnected layers including spatial convolution layers, half-wave rectification layers, spatial pooling layers, normalization layers, and fully connected layers. While this method attains outstanding classification performance, it also suffers from large testing complexity, particularly due to the first fully connected layer, as well as large training complexity, since all the coefficients in the pipeline are learned in a supervised manner and require lots of training images. To address this latter issue, \cite{Oquaba} proposed to use DCNN models pre-trained on the Imagenet dataset (consisting of many million images) and then transfer all but the last layer of this pre-trained DCNN to a new target dataset, where two new adaptation layers are learned. This reduces training time and the amount of required training data, but the training data needs to be annotated with bounding box information. The fact that the method works on a per-patch basis further increases the testing complexity relative to standard DCNNs.

Several approaches exist that, like ours, attempt to bridge the classical approach and the DCNN approach using hybrid mixes. Inspired by the popularity of DCNNs, Simonyan \textit{et al.} \cite{simonyan2013deep} proposed to incorporate the deep aspect of DCNNs into traditional SIFT/FV schemes by stacking multiple layers of FV aggregators, with each layer operating on successively coarser overlapping spatial cells. Sydorov \textit{et al.} \cite{sydorovdeep} instead proposed viewing the standard FV aggregator as a deep architecture, substituting the unsupervised GMM parameters of the FV aggregator by supervised versions. 

While these methods adopted only the deep aspect of DCNNs, our goal is to combine the advantages of both approaches (DCNNs and classical aggregators) using hybrid mixes of both pipelines. We do this by treating the output of the pre-trained intermediate layers of the DCNN architecture as local image descriptors, which we aggregate using standard aggregators such as BoW or FV. There is no need to carry out costly tuning of the DCNN adaptation layers \cite{Oquaba} to the target dataset, as both BoW and FV rely on unsupervised learning. The closest related method in the literature is that of Gong \textit{et al.} \cite{gong2014multi}, who propose using the output of the previous-to-last fully connected layer as a local descriptor, computing this descriptor on multi-scale dense patches subsequently aggregated using VLAD on a per-scale basis. This approach is very complex because, contrary to our approach, one needs to compute the full DCNN pipeline not only on the original image but also on a large number of multi-scale patches and further apply two levels of PCA dimensionality reduction. 


The remainder of this paper is organized as follows: In \sxnref{background}, we describe the two classical aggregators (BoW and FV) that we use in our experiments, as well as the DCNN architecture. In \sxnref{proposed}, we describe our hybrid image feature extraction pipeline. We evaluate our proposed method in \sxnref{results} and provide concluding remarks in \sxnref{conclusion}.

\section{Background}\label{background}
In this section we present an overview of two classical local descriptor aggregation methods: the BoW aggregator \cite{Sivic2003,Lazebnika,Bosch} and the FV aggregator \cite{Perronnin2010}. Up until recently, such aggregation schemes together with SVM classifiers were the reference in image classification \cite{Chatfield2011}.  We then present an overview of the new state-of-the art DCNN image classification pipeline \cite{Krizhevsky2012}. 

\subsection{Image Classification using Local Descriptor Aggregators}
The classical image classification procedure consists of first mapping images to a fixed-dimensional image feature space where linear classifiers are computed using SVMs. The image feature construction process operates by aggregating the local descriptors extracted from the image in question, $\v f: \{\v x_k \in \mathds R^d\}_k \mapsto \mathds R^D$, where the $\v x^k$ are the local descriptors of the image.

The Bag-of-Words (BoW) aggregator offers one such way to map local descriptors to image features. A training set of local descriptors $\s T$ from a representative set of images is first used to build a codebook $\m C = [\v c_j]_j$ using $K$-means. Letting $\s C_j$ denote the Voronoi cell for codeword $\v c_j$, the BoW aggregated image feature is the relative frequency of occurrence of local descriptors in the Voronoi cells: 
\begin{equation}\label{eqn:BoW}
\v f = \left [ {\# \left ( \{\v x_k, \v x_k \in \s C_j\}_k \right )} / {\# \left (\{\v x_k\}_k \right )} \right ]_j,
\end{equation}
where we let $\#$ denote set cardinality. The BoW encoder offers an intuitive image feature and enjoys a low computational cost that can be important in user-in-the-loop applications such as \cite{Parkhi}. 

A more recent image feature, the Fisher vector, offers an important gain in image classification performance \cite{Chatfield2011}. The Fisher encoder requires that a training set of local descriptors $\s T$ be used to learn a GMM model $\s G=\{\beta_k,\m \Sigma_k, \v c_k\}_k$ with $k$-th mixture component having prior weight $\beta_k$, covariance matrix (assumed diagonal) $\m \Sigma_k$ and mean vector $\v c_k$. The first order Fisher vector for a given image can then be computed as follows:
\begin{equation}
\label{eqn:Fisher}
\v f = \left [ \frac{1}{M}\sum_{k=1}^M \frac{p(j|\v x_k)}{\sqrt{\beta_j}} \m \Sigma_j^{-1} \left ( \v x_k - \v c_j \right ) \right ]_j.
\end{equation}

Both the BoW and Fisher aggregators are built from unsupervised models for the distribution of local descriptors, with supervision coming into play only at the classifier learning stage. Deep CNNs instead construct a fully supervised image-to-classification score pipeline.

\subsection{Deep Convolutional Neural Networks (DCNNs)}
\begin{figure} 
\includegraphics[angle=90, width=\columnwidth]{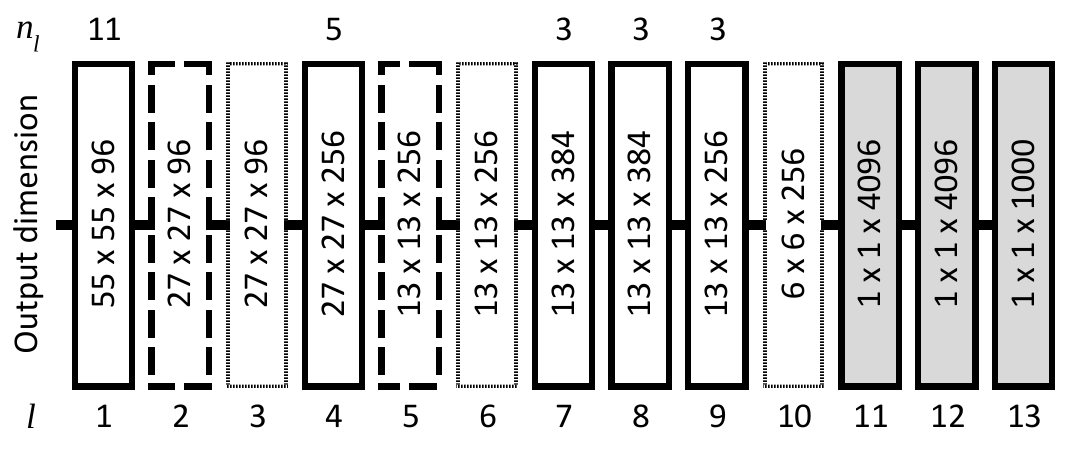}
\caption{Architecture of the Deep-CNN pipeline of \cite{Krizhevsky2012} trained on ImageNet 2012 and used in this paper. Each layer, represented by a box, is labeled with the size $R_l \times C_l \times K_l$ of its output in \eqref{eqn:cnn as local descs}. The $K_l$ kernels at layer $l$ have dimension $n_l \times n_l \times K_{l-1}$. The layer index $l$ (respectively, kernel spatial dimension $n_l$) is indicated below (above) the box for each layer. The input image is assumed normalized to size $224 \times 224 \times 3$, and $4\times$ downsampling is applied during the first layer. \emph{Dark-lined boxes:} convolutional layers; \emph{dash-lined boxes:} normalization layers; \emph{light-lined boxes:} max-pooling layers; \emph{grayed-in boxes:} fully-connected layers.}
\label{fig:DCNN}
\end{figure}
Deep Convolutional Neural Networks have established an overwhelming presence in image classification starting with the 2012 ImageNet Large Scale Visual Recognition Challenge (ILSVRC) \cite{Krizhevsky2012}. The performance gap of DCNNs relative to the second entry in that year's competition (and relative to SIFT-based Fisher aggregation schemes \cite{Sanchez2011}) is in excess of 10 percentage points in absolute improvement of top-5 error rate. 

In \figref{fig:DCNN} we illustrate the deep DCNN processing pipeline of \cite{Krizhevsky2012}. It consists of convolutional layers, max-pooling layers, normalization layers and fully connected layers. At any given layer $l$, the layer's output data is an $R_l \times C_l \times K_l$ array 
\begin{equation} \label{eqn:cnn as local descs}
[ \v x^l_{ij} \in \mathds R^{K_l} ]_{i=1,\ldots,R_l;\; j=1,\ldots,C_l},
\end{equation}
that is the input to the next layer, with the input to layer $l=1$ being an RGB image of size $R_0 \times C_0$ and $K_0=3$ color channels. 

The  \emph{convolutional layers} ($l=1,4,7-9$) first compute the spatial convolution of the input with $K_{l}$ kernels of size $n_l \times n_l \times K_{l-1} $ and then apply entry-wise Rectified Linear Units (ReLUs)  $\max(0,z)$. 
The \emph{normalization layers} ($l=2,5$) normalize each $\v x \in \{\v x^{l-1}_{ij}\}_{ij}$ at the input using what can be seen as a generalization of the $l_2$ norm consisting of dividing each entry $x_m$ of $\v x$ by $ (2 + 10^{-4} \sum_{n \in \s I_m} x_n^2)^{0.75}$. The summation indices $\s I_m$ are taken to be the $m$-th sliding window over the indices of all entries.
The \emph{max-pooling layers} ($l=3,6,10$) carry out per-kernel spatial max-pooling by taking the maximum value from each spatial bin of size $3 \times 3$ spaced every $2$ pixels.

The \emph{fully connected layers} ($l=11-13$) can be seen as convolutional layers with kernels having the same size as the layer's input data. The last layer ($l=13$) uses a softmax non-linearity instead of the ReLU non-linearity used in other layers and acts as a multi-class classifier, having as many outputs as there are classes targeted by the system.


\subsection{Transfer learning using DCNNs} The architecture in \figref{fig:DCNN} contains more than 60 million parameters and training it can be a daunting task requiring expensive hardware, large annotated training sets (ImageNet 2012 contains 15 million images and 22,000 classes) and training strategies including memory management schemes, data augmentation and specialized regularization methods. Moreover, extending the architecture to new classes would potentially require re-training the entire structure, as the full architecture is learned for a specific set of target classes.

To address this last difficulty, Oquab \etal \cite{Oquaba} use transfer learning to apply the architecture in \figref{fig:DCNN} to new classes while incurring reduced training overhead. Their approach consists of substituting only the last fully-connected classification layer by two learned adaptation layers, a fully-connected ReLU layer with $4096$ neurons followed by a fully-connected softmax classification layer with as many neurons as target classes. The first $12$ layers are transferred from the net in \figref{fig:DCNN} (learned from ImageNet 2012 data), and only the new adaptation layers are learned using training data for the new set of target classes  (\eg those of the Pascal VOC 2007 test bench). 

While their approach reduces the training overhead and required training set size, training the adaptation layers still requires non trivial complexity as these contain a large number of parameters (more than 16 million ). To obtain an adequately large training set from Pascal VOC 2007 data, they derive a patch-based training set, labeling every patch according to its intersection with the provided object bounding boxes. Their approach thus operates on a per-patch classification basis, and the overall class score is obtained by summing this per-patch scores over the entire image for each class. This brings the important benefit of also providing the object localization, but it requires laborious bounding-box annotations on the training set and costly training of millions of parameters.


\section{A hybrid DCNN/Aggregator feature}\label{proposed}
Inspired by the transfer learning approach of \cite{Oquaba}, in this section we propose a new hybrid feature that combines parts of the DCNN architecture in \figref{fig:DCNN} trained on ImageNet 2012 with the unsupervised BoW or Fisher local descriptor aggregation schemes in \eqref{eqn:BoW} and \eqref{eqn:Fisher}. The resulting feature is used with one-vs-all linear SVM classifiers and hence new classes can be added with little training overhead and without the need for costly object bounding box annotations.

\subsection{Per-layer aggregation of DCNN local descriptors}
\label{hybrid}
Our hybrid scheme is based on the observation that the vectors $\v x^l_{ij}$ in \eqref{eqn:cnn as local descs} comprising the output of layers $l=1,\ldots,10$ in \figref{fig:DCNN} (\ie all layers except fully-connected layers) can be treated as densely extracted local descriptors. We will hence build one aggregated feature $\v f_l$ for each layer $l$ (or a subset of layers $l \in \s L$) and concatenate all the resulting aggregated layer features to form a single image feature 
\begin{equation} \label{eqn:hybrid}
\v f = [\v f_l^T]_{l\in \s L}^T.
\end{equation}
Using only a subset of layers $\s L \subseteq \{1,\ldots,10\}$ allows us to control training, testing and storage complexity and further serves as a means of regularization.

\subsection{Training per-layer aggregators}
In order to train the per-layer aggregators adapted to the DCNN layers, we take each image from a representative set of training images and extract from it all vectors $\v x^l_{ij}$ for $l=1,\ldots,10$. We then group all the resulting local descriptors $\v x^l_{ij}$ for each layer $l$ to form a training set $\s T^l$ for the $l$-th layer. Each training set $\s T^l$ of local descriptors is then used to train a codebook $\m C^l$ for layer $l$ using $K$-means when using BoW aggregators. Likewise, a GMM model $\s G^l$ is learned for the $l$-th layer when using Fisher aggregators.

\subsection{Extensions based on classic approaches}
Our proposed approach shares similarities with several existing approaches and we now discuss these and related extensions.

One first observation is that the spatial support (relative to the original image) used to compute the $\v x_{ij}^l$ is of size $11$ (in each spatial dimension) for the first layer and grows by $4 \times 2 \cdot (n_a-1)$ for each convolutional layer $ 1 < a \leq l$, yielding possible supports of size $11,43,59$ and $75$. Dense approaches likewise compute local descriptors from supports of varying size ($16,24,32,40$) by means of multi-resolution spatial grids \cite{Chatfield2011}, but all descriptors for all supports are pooled together (for the benefit of scale invariance) and used to form a single aggregated image feature. A similar pooling approach could be used for DCNN local descriptors $\v x_{ij}^l \in \mathds R^{K_l}$ by first mapping all layers to a common dimensionality via, \eg PCA or discriminative dimensionality reduction.

The layer feature concatenation scheme \eqref{eqn:hybrid} that we use instead is reminiscent of spatial pyramid matching \cite{Lazebnika,Bosch}, where one feature $\v g_{c}$ is computed for each spatial cell $c=1,\ldots,8$ and these are subsequently concatenated. Our concatenated image features $\v f_l$ are instead computed from high-dimensional filtered versions of the image, and indeed this approach can be combined with SPM to produce per-spatial-cell layer features $\v f_{lc}$.

Other standard successful approaches can also be combined with our proposed hybrid DCNN/aggregator features, including power normalization of the $\v x_{ij}^l$ \cite{Arandjelovic2012}, application of an explicit Hellinger kernel-map to our hybrid feature \cite{Chatfield2011} and late fusion with other feature channels.
Alternate aggregation schemes such as VLAD or triangulation embedding \cite{Delhumeau2013,Zisserman2014} can also be used, but we chose BoW for its low computational cost and Fisher given that is the best performing aggregator in classification.

\section{Results}\label{results}
In this section we validate our proposed hybrid DCNN/aggregator feature using the publicly available Pascal VOC 2007 dataset \cite{pascal2007}. This dataset consists of 9163 images representing 20 visual categories and split into training, validation and test sets. We use the standard mean Average Precision (mAP) measure computed over the test set as a performance metric.


\subsection{Impact of layer subset $\s L$}
\label{sec:cases}
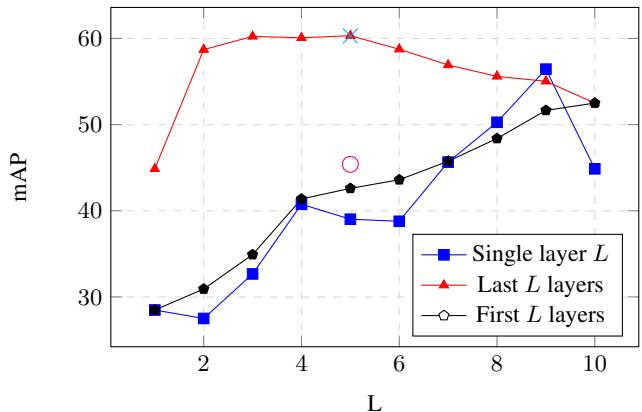
\begin{figure}[!ht]
\begin{tikzpicture}

\begin{axis}[       
       grid=major,
       grid style={dashed,gray!30},
       xlabel=L,
       ylabel=mAP,
       legend pos={south east},
       legend style={font=\small},
       plots
       ]
\addplot [color=blue,mark=square*] coordinates
{ (1,28.48)
  (2,27.50)
  (3,32.67)
  (4,40.76)
  (5,39.02)
  (6,38.78)
  (7,45.65)
  (8,50.27)
  (9,56.44)
  (10,44.87)};
\addlegendentry{Single layer $L$}
\addplot [color=red,mark=triangle*] coordinates
{ (1,44.87)
  (2,58.71)
  (3,60.24)
  (4,60.08)
  (5,60.32)
  (6,58.75)
  (7,56.92)
  (8,55.60)
  (9,55.03)
  (10,52.50)};

\addlegendentry{Last $L$ layers}

\addplot [mark=pentagon*] coordinates
{ (1,28.48)
  (2,30.92)
  (3,34.91)
  (4,41.37)
  (5,42.60)
  (6,43.60)
  (7,45.75)
  (8,48.39)
  (9,51.65)
  (10,52.50)};
\addlegendentry{First $L$ layers}

\addplot [color=cyan,only marks,mark=x,mark options={scale=2}] coordinates { (5,60.30)};
\addplot [color=magenta,only marks,mark=o,mark options={scale=1.5}] coordinates{(5,45.39)};

\end{axis}
\end{tikzpicture}

\caption{The mAP is plotted for hybrid features built using a single layer, the last L layers and the first L layers (excluding fully connected layers 11-13), for codebook size 500. Baseline results for BoW and FV are displayed using $\circ$ and $\times$ markers.}
\label{fig:layer}

\end{figure}

In \figref{fig:layer} we evaluate the impact on performance of the layer subset $\s L$ in \eqref{eqn:hybrid} used to build hybrid features. We consider three strategies for selecting $\s L$: using a single layer, $\s L=\{L\}$, using the first $L$ layers, $\s L=\{1,\ldots,L\}$, and using the last $L$ layers, $\s L=\{10,9,\ldots,10-L+1\}$. As seen in \figref{fig:layer}, the results for the single-layer strategy indicate that layers further down the pipeline are more informative (although the curve is not monotonic). Indeed the best strategy overall consists of using the last $5$ layers (and using only $3$ layers results in marginal performance decrease). The resulting hybrid feature performs substantially better than BoW+SPM with $4,000$ codewords and performs similar to FV+SPM with $256$ mixture components \cite{perronnin}, despite being 150 times smaller.

\subsection{Impact of codebook size}
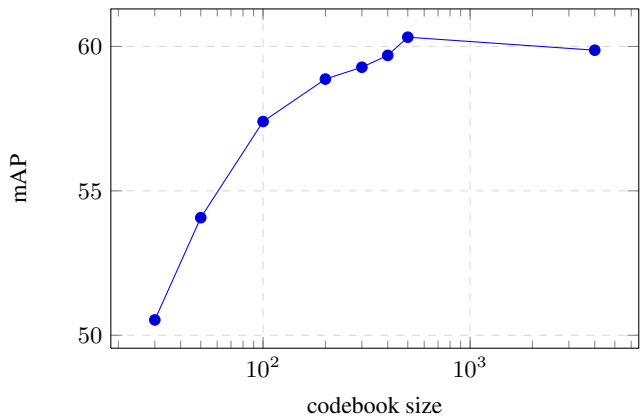
\begin{figure}[!ht]
\begin{tikzpicture}
\begin{axis}[       
       grid=both,       
       grid style={dashed,gray!30},
       xlabel=codebook size,
       ylabel=mAP,
       legend style={at={(0.35,-0.20)},anchor=north},
       xmode=log,
       plots
       ]
\addplot coordinates
{ 
  
  (30,50.53)
  (50,54.07)
  (100,57.40)
  (200,58.87)
  (300,59.28)
  (400,59.69)
  (500,60.32)
  (4000,59.87)
 
};

\end{axis}
\end{tikzpicture}

\caption{The mAP vs the codebook size in log scale when using last L=5 layers.}
\label{fig:codebook}
\end{figure}

In \figref{fig:codebook} we evaluate the impact on performance of varying the codebook size when using hybrid DCNN/BoW  features built from the last $5$ layers. A codebook of size $500$ yields the best performance. And even with a codebook size of $30$, which amounts to a feature vector size of 150, our method outperforms BoW + SPM.

\begin{figure}[!ht]
\begin{tikzpicture}
\begin{axis}[       
       grid=major,
       grid style={dashed,gray!30},
       xlabel=L,
       ylabel=mAP,
       legend style={at={(0,-0.40)},anchor=west},
       plots
       ]
\addplot  [color=red,mark=triangle*] coordinates
{ 
  (1,70.92)
  (2,73.78)
  (3,73.81)
  (4,73.89)
  (5,73.90)
  (6,73.90)
  (7,73.90)
  (8,73.90)
  (9,73.90)
  (10,72.54)
  (11,72.50)
  (12,72.31)
  (13,71.22)
 
};

\end{axis}
\end{tikzpicture}
\caption{Using last L layers  from \figref{fig:DCNN}. Here we include the fully connected layers 11-13.}
\label{fig:lastn}
\end{figure}
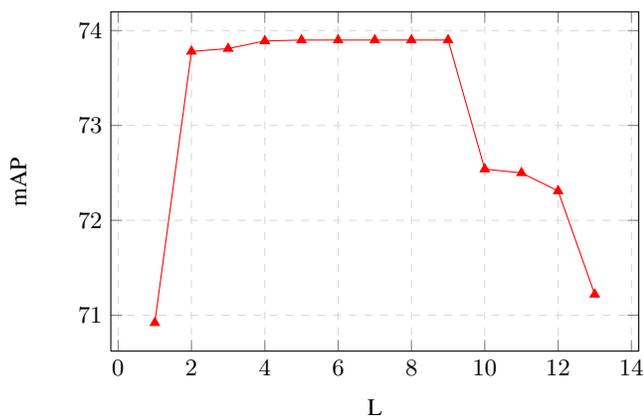

\begin{table}[!ht]
\centering
\begin{tabular}{|c|c|}
\hline
method& Training time+resource \\
\hline
PRE1000C \cite{Oquaba}& $\approx$ 1 day (GeForce GTX Titan GPU)\\
Hybrid DCNN/BoW, N=500& $\approx$ 1hr + 5min (8 core CPU)\\
\hline
\end{tabular}
\caption{Table illustrating training time for 500 codebook size and when using the last 5 layers. Training times are for the unsupervised learning part with and without supervised learning of linear SVM classifiers for all Pascal VOC 2007 classes. This is compared to the training time taken by the method \cite{Oquaba}.}
\label{complexity}
\end{table}

\subsection{Comparison to other approaches}
In Table \ref{compare} we compare our results with some of the best results reported in the literature. We include results for hybrid features built using FV aggregators with $64$ mixture components. Despite the established superiority of FV aggregation over BoW aggregation, the FV-based hybrid features perform poorly relative to BoW-based hybrid features. We believe that this is due to the small number of local descriptors in DCNN layers, as this makes the vector-averaging process in \eqref{eqn:Fisher} statistically noisy.

The best performing system in \tblref{compare} is PRE1000C \cite{Oquaba}. Their approach consists of substituting layer $13$ in \figref{fig:DCNN} by two adaptation layers trained on Pascal VOC. As is the case for DCNN pipelines, this training procedure is time consuming and requires expensive GPU cards, as illustrated in  \tblref{complexity}. Furthermore, at testing time, their approach requires applying the full $13$-layer DCNN pipeline to each of 500 patches from an image, increasing testing complexity considerably. Our approach requires a single DCNN pipeline pass over the non fully-connected layers, resulting in dramatically lower testing time, as the DCNN complexity is largely concentrated in the first fully-connected layer. 

The same complexity problem is incurred by the feature construction scheme of \cite{gong2014multi}, where the authors propose using the output of DCNN layer $13$ as a local descriptor computed on multi-scale dense image patches. Inspired by this approach, we further consider stacking the output of the fully connected layers ($11$, $12$, and $13$) to our hybrid DCNN/aggregator feature. We illustrate the results of this approach in \figref{fig:lastn}, where the non-fully connected layers are processed according to \eqref{eqn:hybrid}, and the fully-connected layers are concatenated without any processing. Note that using the $3$ fully connected layers and the last non-fully connected layer results in performance close to $74$ mAP points. This compares very well to the performance of $77.73$ of PRE1000C in \tblref{compare}, particularly considering the drastic difference in training time and testing time.


\begin{table}[!ht]
\centering
\begin{tabular}{c|c|c}
\hline
method& feature dimension & mAP\\
\hline
BoW + SPM, N=4000 \cite{Chatfield2011} & 32000 & 45.39\\
FV (SIFT) \cite{perronnin}& 262144 & 58.3\\
FV (SIFT + color) \cite{perronnin} & 262144 & 60.3\\
PRE1000C \cite{Oquaba} &&77.73\\
\hline
Hybrid DCNN/FV, m=64 &81920&54.56\\
Hybrid DCNN/BoW, N=30 & 150&50.53\\
Hybrid DCNN/BoW, N=500 & 2500&60.32\\

\hline
\end{tabular}
\caption{Comparison of our results (using last 5 layers) with the state-of-the-art (N represents the codebook size in BoW).}
\label{compare}
\end{table}
\section{Conclusion}\label{conclusion}
In this work, we proposed a hybrid Deep Convolutional Neural Network (DCNN) / Bag-of-Words (BoW) image feature extraction approach. Treating the output of intermediate layers of a pre-trained DCNN as local descriptors allowed us to use an unsupervised  Bag-of-Words aggregator to obtain an image feature that outperforms standard aggregators based on local descriptors substantially on the Pascal VOC 2007 benchmark. Appending the output of the fully-connected layers to our hybrid feature further improves the performance of our approach, making it competive with DCNNs variants adapted to Pascal VOC 2007, and at a fraction of the training and testing cost.



\bibliographystyle{IEEEbib}
\bibliography{collection1}

\end{document}

%% file: notation.tex
\usepackage{amsmath}
\usepackage{amssymb}
\usepackage{dsfont}

\newcommand{\ie}[0]{\textit{i.e.},~}
\newcommand{\eg}[0]{\textit{eg.},~}

\newcommand{\etal}[0]{\textit{et al.}~}

\newcommand{\s}[1]{\ensuremath{\mathcal{#1}}} 
\renewcommand{\v}[1]{\ensuremath{\mathbf{#1}}} 
\newcommand{\m}[1]{\ensuremath{\mathbf{#1}}} 

\newcommand{\figref}[1]{Fig. \ref{#1}}
\newcommand{\sxnref}[1]{Section \ref{#1}}
\newcommand{\tblref}[1]{Table \ref{#1}}